\documentclass[letterpaper, 10 pt, conference]{IEEEtran}

\IEEEoverridecommandlockouts

\usepackage{cite}
\usepackage{textcomp}
\usepackage[hyphens]{url}
\usepackage[utf8]{inputenc}
\usepackage[T1]{fontenc}
\usepackage{makecell}
\usepackage{multirow}
\usepackage{graphicx}
\usepackage{subcaption}
\usepackage{pgfplots}
\usepackage{amssymb,amsmath}
\usepackage[tight-spacing=true]{siunitx}
\usepackage{float}
\usepackage[colorlinks,bookmarksopen,bookmarksnumbered,citecolor=red,urlcolor=red]{hyperref}
\usepackage{adjustbox}
\usepackage{tabularx}
\usepackage{booktabs}
\newcommand{\ra}[1]{\renewcommand{\arraystretch}{#1}}
\newcommand{\minitab}[2][l]{\begin{tabular}{#1}#2\end{tabular}}
\newcolumntype{Y}{>{\raggedright\arraybackslash}X}
\newcolumntype{b}{Y}
\newcolumntype{m}{>{\hsize=.55\hsize}Y}
\newcolumntype{s}{>{\hsize=.15\hsize}Y}
\newcolumntype{v}{>{\hsize=.65\hsize}Y}

\newcommand\BibTeX{{\rmfamily B\kern-.05em \textsc{i\kern-.025em b}\kern-.08em
T\kern-.1667em\lower.7ex\hbox{E}\kern-.125emX}}

\begin{document}
	

\title{CADDY Underwater Stereo-Vision Dataset for Human-Robot Interaction (HRI) in the Context of Diver Activities}


\author{Arturo Gomez Chavez\textsuperscript{1}, Andrea Ranieri\textsuperscript{2}, Davide Chiarella\textsuperscript{2}, Enrica Zereik\textsuperscript{2},\\Anja Babić\textsuperscript{3}, Andreas Birk\textsuperscript{1}
	
\thanks{\textsuperscript{1} Robotics Group of the Computer Science \& Electrical Engineering Department.Jacobs University Bremen gGmbH, Germany.  \texttt{\{a.gomezchavez,\allowbreak a.birk\}@jacobs-university.de}.}
\thanks{\textsuperscript{2} Institute of Intelligent Systems and Automation. National Research Council of Italy (CNR).}
\thanks{\textsuperscript{3} Faculty of Electrical Engineering and Computing. University of Zagreb, Croatia.}
}


\maketitle

\begin{abstract}
	In this article we present a novel underwater dataset collected from several field trials within the EU FP7 project ``Cognitive autonomous diving buddy (CADDY)'', where an Autonomous Underwater Vehicle (AUV) was used to interact with divers and monitor their activities. 
	To our knowledge, this is one of the first efforts to collect a large dataset in underwater environments targeting object classification, segmentation and human pose estimation tasks. 
	The first part of the dataset contains stereo camera recordings ($\approx10K$) of divers performing hand gestures to communicate and interact with an AUV in different environmental conditions.
	These gestures samples serve to test the robustness of object detection and classification algorithms against underwater image distortions i.e., color attenuation and light backscatter.
	The second part includes stereo footage ($\approx12.7K$) of divers free-swimming in front of the AUV, along with synchronized IMUs  measurements located throughout the diver's suit (\textit{DiverNet}) which serve as ground-truth for human pose and tracking methods.
	In both cases, these rectified images allow investigation of 3D representation and reasoning pipelines from low-texture targets commonly present in underwater scenarios. 
	In this paper we describe our recording platform, sensor calibration procedure plus the data format and the utilities provided to use the dataset. 
\end{abstract}


\section{Introduction}
\label{sec:introduction}

From the robotics perspective, underwater environments present numerous technological challenges for communication, navigation, image processing and other areas. 
There are unique sensors problems due to the electromagnetic waves being attenuated very strongly: no GPS-based localization, no radio communication and only limited possibility of using visible light. 
Acoustics sensors are mostly used, but they offer low bandwidth and high latency transmissions. 
It is no surprise that most of the applications such as biological sample acquisition, archaeological site exploration, industrial panel manipulation, etc; still require human intervention for their successful completion.

For this reason, the EU FP7 CADDY project focused on diver-robot cooperation; 
where an AUV monitored divers activities, while communicating with them and performing multiple tasks on command \cite{Miskovic2017_caddy-y3}. 
Along the duration of this project, a unique set of data was recorded covering diver gesture recognition based on the CADDIAN sign language \cite{Chiarella15_CADDIAN}, and diver pose estimation using stereo-images and inertial sensor measurements from a specifically made diver's suit, called \textit{DiverNet}~\cite{Goodfellow2015_divernet}. 

\begin{figure}[!b]
	\centering
	\captionsetup{justification=justified}
	\begin{subfigure}{\linewidth}
		\centering
		\includegraphics[width=.71\linewidth]{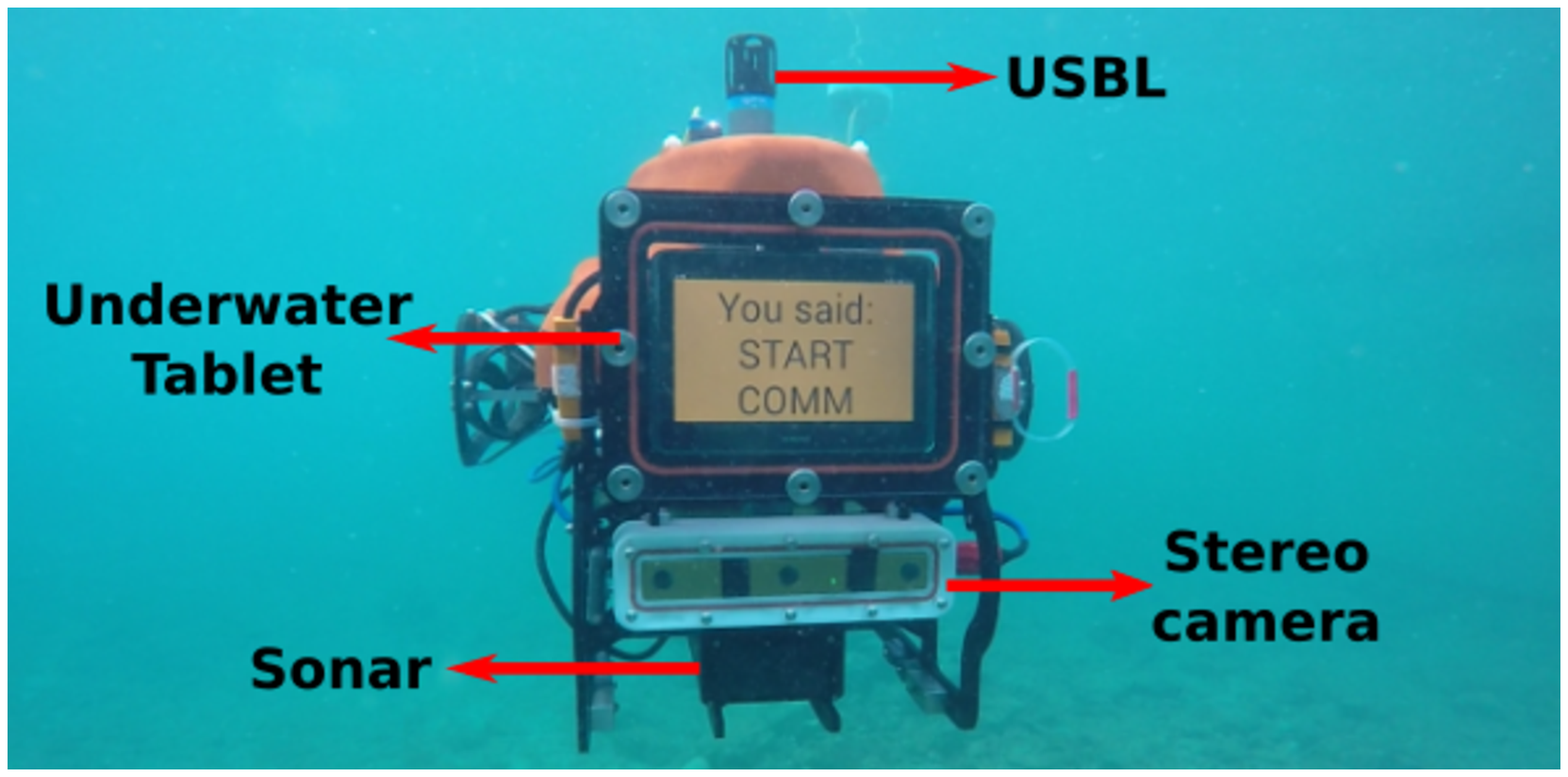}
	\end{subfigure}
	\begin{subfigure}{\linewidth}
		\centering
		\includegraphics[width=.71\linewidth]{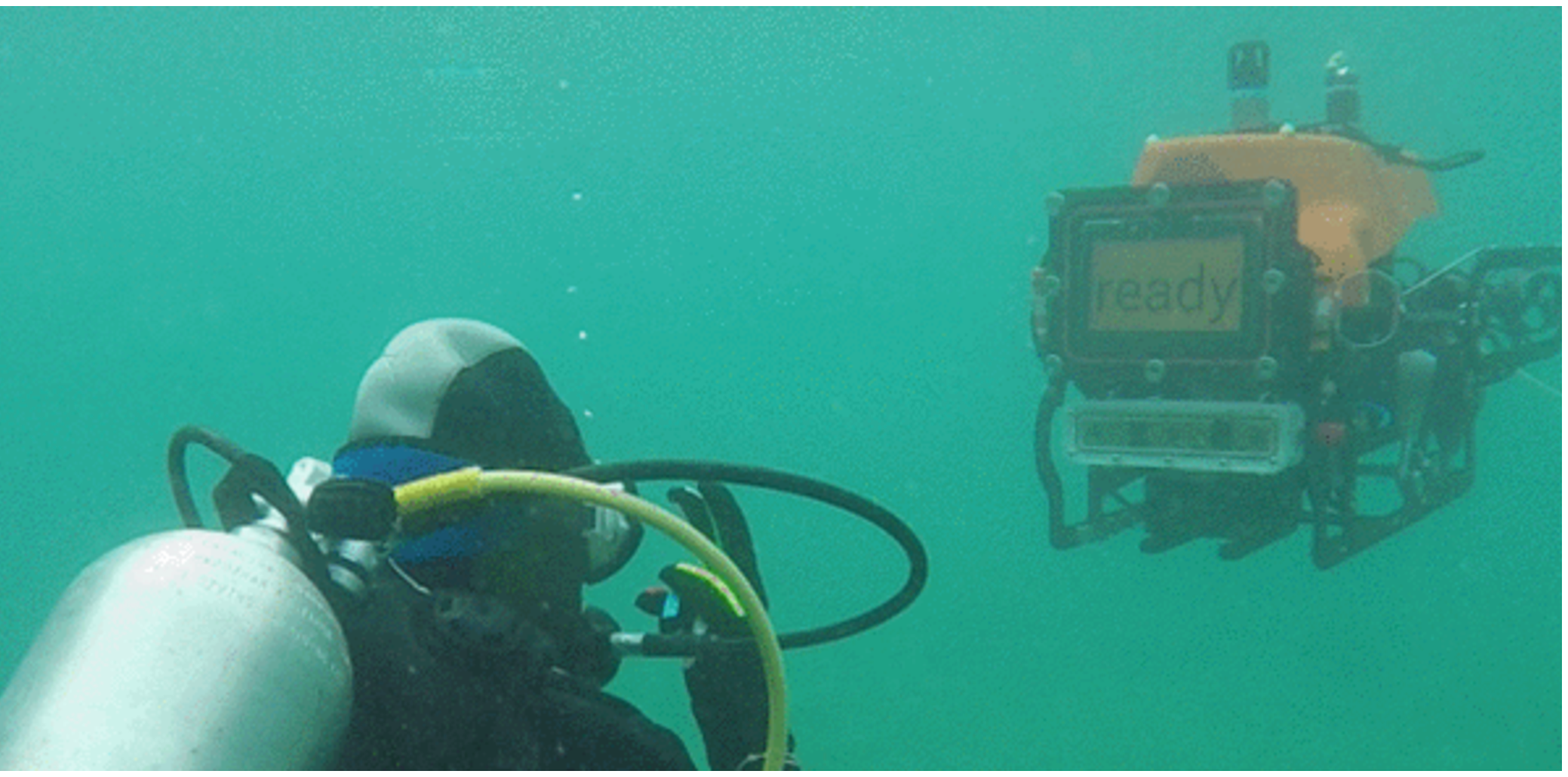}
	\end{subfigure}
	\caption{(Top) BUDDY-AUV equipped with sensors to monitor the diver - Blueprint Subsea X150 USBL, ARIS 3000 Multibeam Sonar, BumbleBeeXB3 Stereo Camera, Underwater Tablet. (Bottom) Diver issuing gesture commands to the AUV, reprinted from \cite{Miskovic2017_caddy-y3}}
	\label{fig:buddy-auv}
\end{figure}

To the best of our knowledge this is the first underwater dataset focusing on human-robot interaction between AUV and diver. 
The data mainly consist of precisely rectified stereo images suited for algorithms using 2D or 3D information, or a fusion of both;
and the number of samples exhibited and their intravariance allow for testing feature-based or deep learning reasoning methods. 

Although there are exhaustive underwater field trials surveys and performance evaluations of object recognition and stereo systems \cite{Garcia2011_turbidImages} \cite{Rizzini2015_uwStereo}; due to the difficulty and cost of underwater data collection, the authors have to benchmark their methods across different application datasets available or use very constrained ones. The work here presented aims to solve this through a sufficiently large and representative dataset.

It is important to note that the use of these data is not limited to recognition and pose estimation tasks; it can serve as a basis for any vision-based algorithm in underwater applications. 
This stems from the fact that data was recorded on different environmental conditions which cause various image distortions unique to underwater scenarios i.e., low contrast, color distortion, haze \cite{Schettini2010_uw_image_processing}, and that cannot be easily replicated from on-land recordings or in simulation. 

\section{Sensor setup}
\label{sec:sensor setup}

\subsection{BUDDY AUV} 
\label{sec:buddy AUV}

For data collection the BUDDY-AUV was specifically designed by the University of Zagreb during the CADDY project \cite{Stilinovic2015_AUV-BUDDY}. 
The vehicle is fully actuated and is equipped with navigation sensors: Doppler Velocity Log (DVL), Ultra Short Baseline (USBL), GPS; and perception sensors: multibeam sonar and Bumblebee XB3 stereo camera. 
In addition, it has a tablet in an underwater housing to enable human-robot interaction capabilities (see Fig.~\ref{fig:buddy-auv}), i.e., output feedback to the diver.  

\subsection{Stereo camera and underwater image rectification}
\label{sec:uw image rectification}
For image collection a Point Grey Bumblebee XB3 color stereo camera was used, model BBX3-13S2C, which provides raw images with $1280 \times 960$ pixels resolution at \SI{16}{Hz}; it has \SI{3.8}{mm} nominal focal length and wide baseline $B=\SI{23.9}{cm}$. 
After rectification, all images are scaled to $640 \times 480$, and these are the dimensions of all the stereo image pairs in this dataset. The camera intrinsic parameter matrix is: $K = \lbrack 710\ 0\ 320;\ 0\ 710\ 240;\ 0\ 0\ 1 \rbrack$.

The camera was enclosed in a watertight housing with a flat glass panel (see Fig.~\ref{fig:buddy-auv}). When using such housing the light is refracted twice: first on the water-glass and then on the glass-water interface. 
These refraction effects cause the image to be distorted; as discussed in \cite{Treibitz2012flat_refractive_geometry}, a camera behind flat glass panel underwater does not possess a single viewpoint and therefore the classic pinhole model is not valid. 
This problem was addressed in \cite{Luczynski2017_Pinax} by proposing a new \textit{Pinax (PINhole-AXial)} camera model that allows for rectification correction by "translating" the image to a rectified, pinhole camera viewpoint. 

\begin{figure}[!b]
	\centering
	\captionsetup{justification=centering}	
	\begin{subfigure}[t]{0.32\linewidth}
		\centering
		\includegraphics[height=2.1cm]{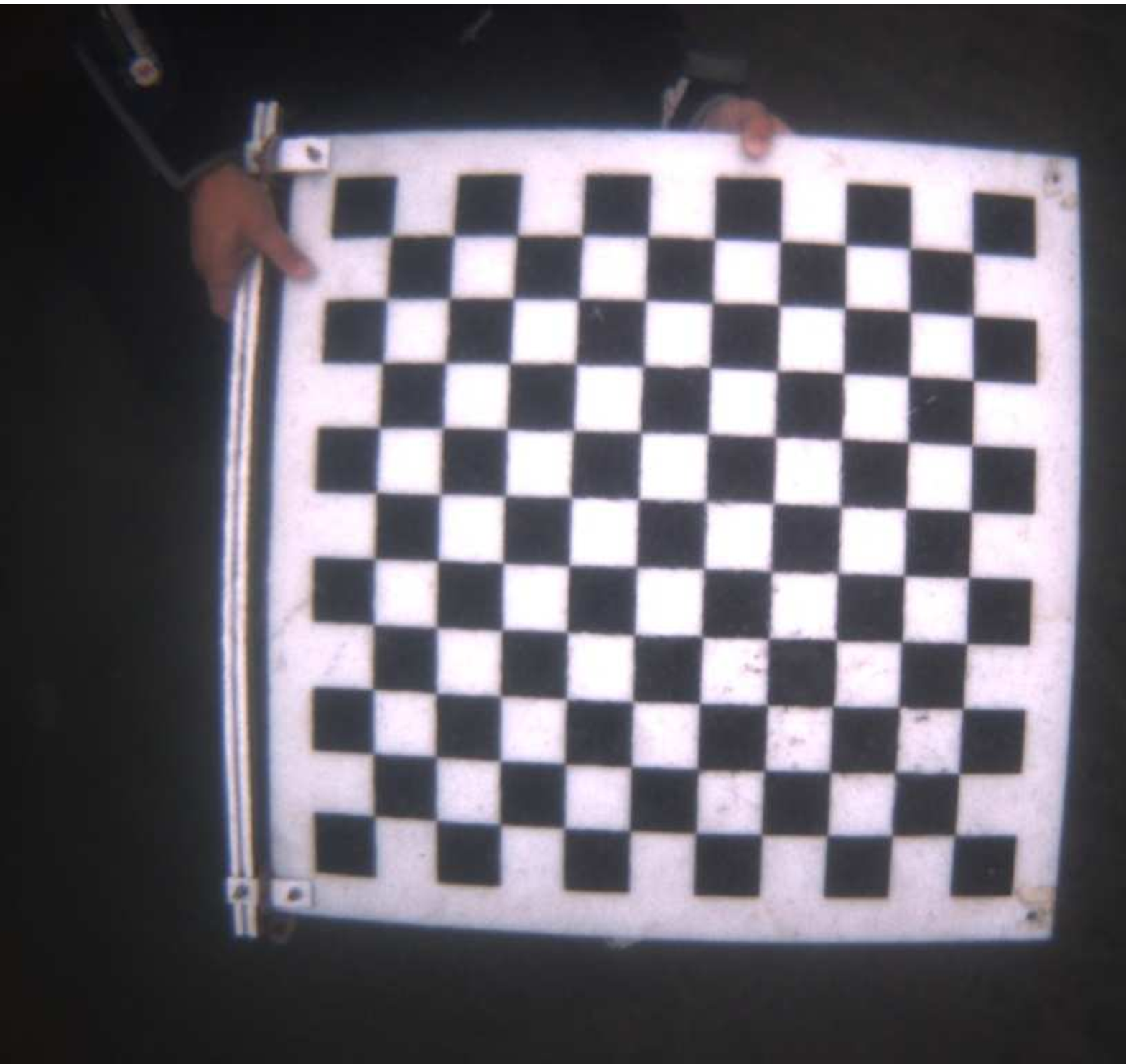}
		\caption{}
	\end{subfigure}
	\begin{subfigure}[t]{0.32\linewidth}
		\centering
		\includegraphics[height=2.1cm]{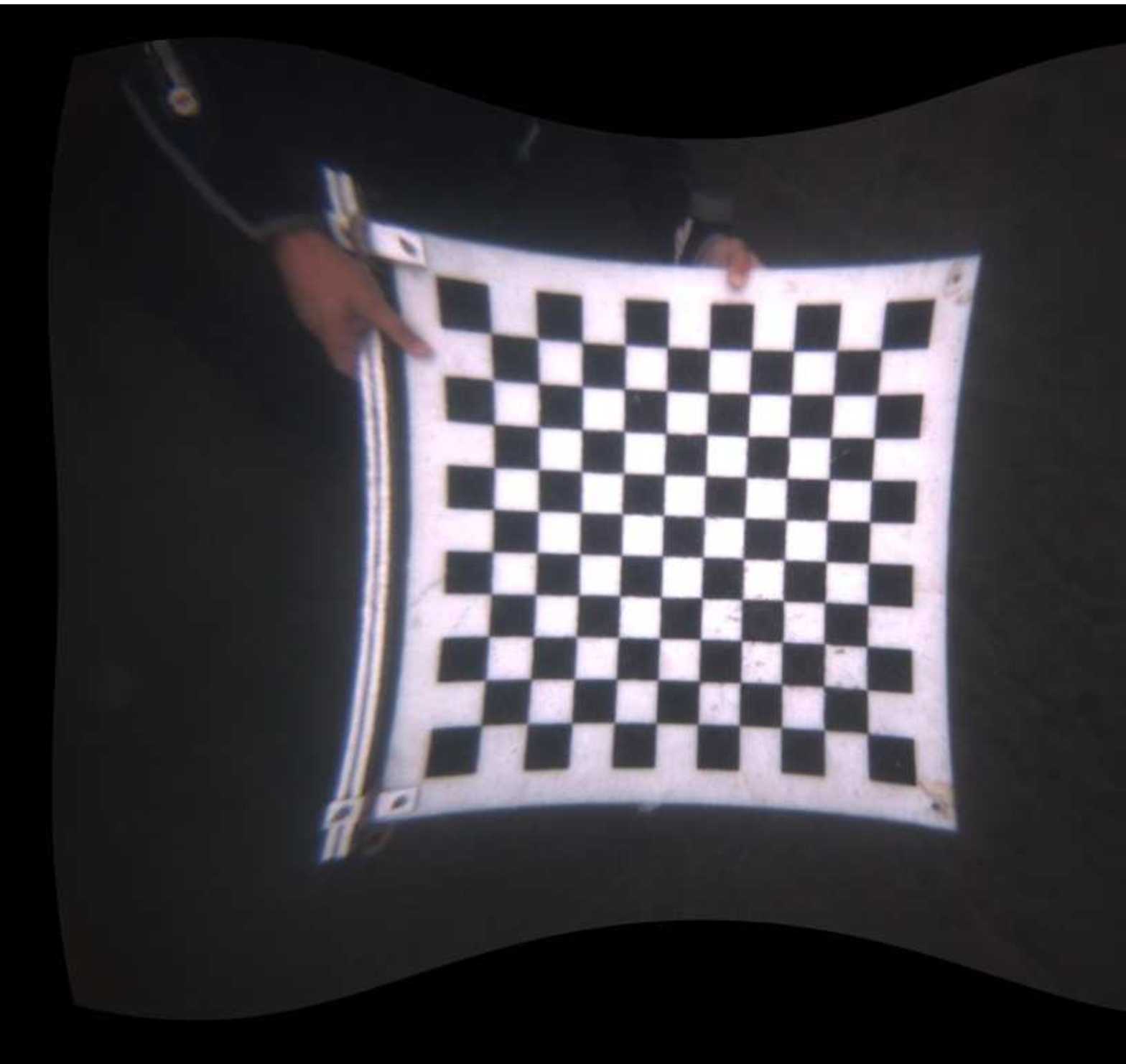}
		\caption{}
	\end{subfigure}
	\begin{subfigure}[t]{0.32\linewidth}
		\centering
		\includegraphics[height=2.1cm]{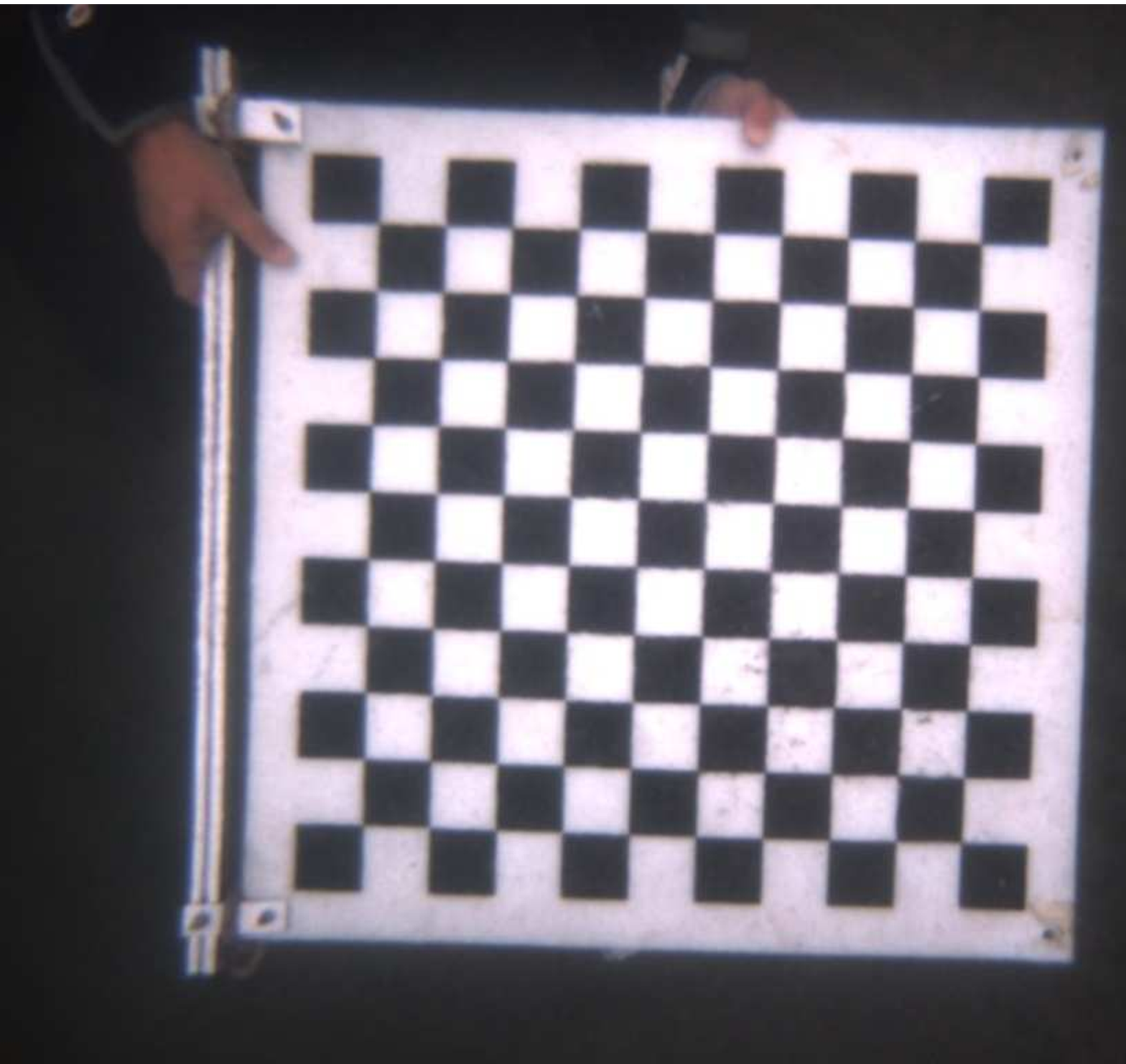}
		\caption{}
	\end{subfigure}
	\captionsetup{justification=justified}
	\caption{(a) Underwater raw image (b) Rectification based on in-air calibration (c) \textit{Pinax} rectification based on water salinity and flat-glass panel thickness.}
	\label{fig:uw_calibration_pattern}
\end{figure}

This method was tested on multiple types of cameras, including Bumblebee XB3, yielding higher quality results than direct underwater calibration i.e., recording a calibration pattern underwater. 
One of the main reasons is that this pattern detection is commonly less accurate when obtained from distorted-raw underwater images, which have low contrast and radial distortions due to magnification artifacts. 
Instead, \textit{Pinax} uses the physical camera model, water salinity and glass thickness to map the air-rectified image to its underwater model. 

Examples of this rectification process are shown in Fig.~\ref{fig:uw_calibration_pattern}; in-air intrinsic calibration was done using the \emph{CamOdCal} software package~\cite{Heng2013_CamOdoCal} with the camera model from \cite{Kannala2006_cameramodel}. The obtained calibration files for the used BumbleBee XB3 instances are provided for the user's inspection, along with the \emph{CamOdCal} and \emph{Pinax} packages in form of a \emph{Docker} container \cite{Merkel2014_docker} which can be used for camera underwater housings with flat-glass panels (\url{git@github.com:jacobs-robotics/uw-calibration-pinax.git}).

\subsection{DiverNet} 
\label{sec:DiverNet}
\begin{figure}[!t]
	\centering
	\captionsetup{justification=centering}	
	\begin{subfigure}[t]{0.48\linewidth}
		\centering
		\includegraphics[height=3.4cm]{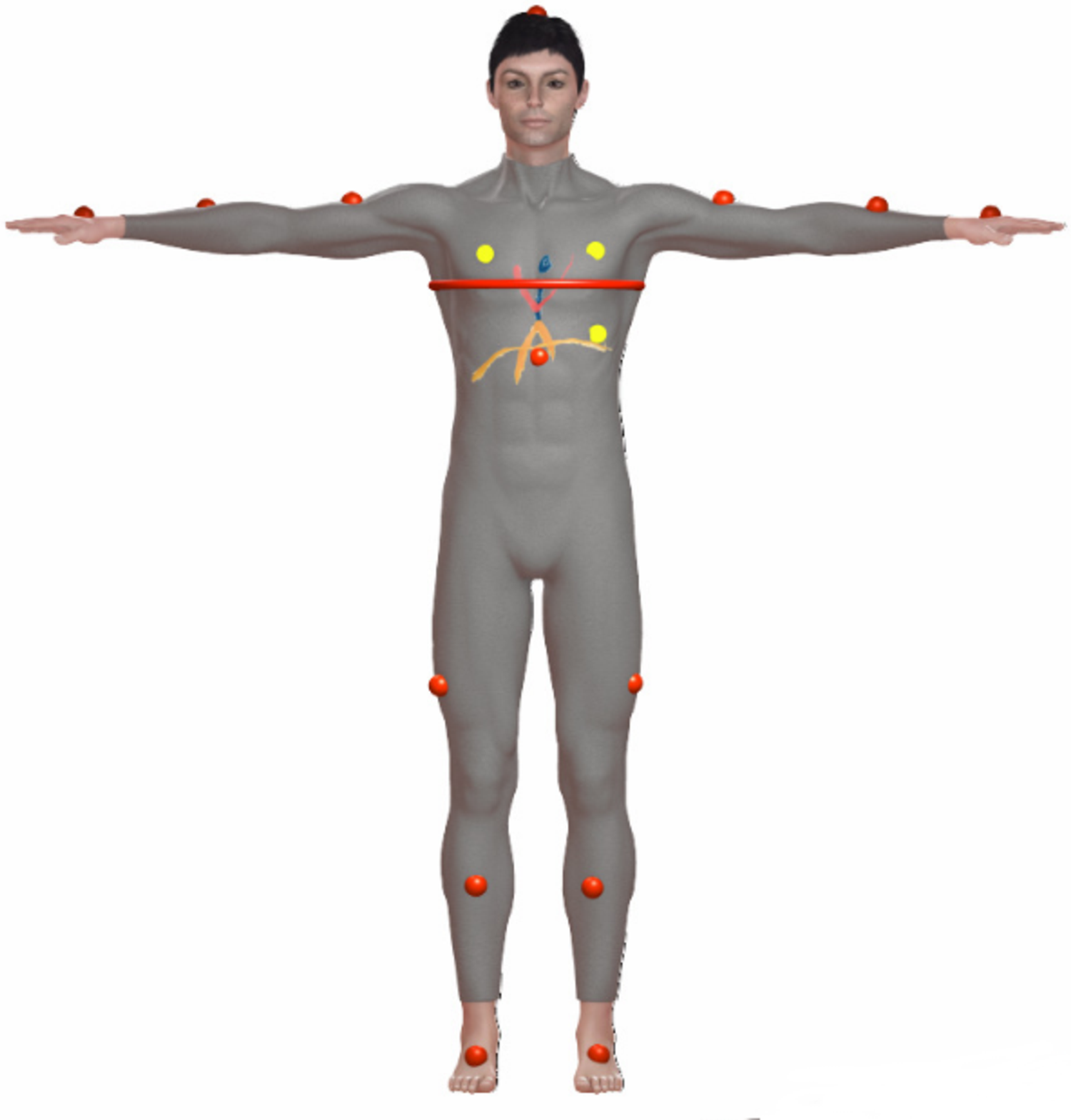}
		\caption{}\label{fig:divernet_sensor_position}
	\end{subfigure}
	\begin{subfigure}[t]{0.48\linewidth}
		\centering
		\includegraphics[height=3.4cm]{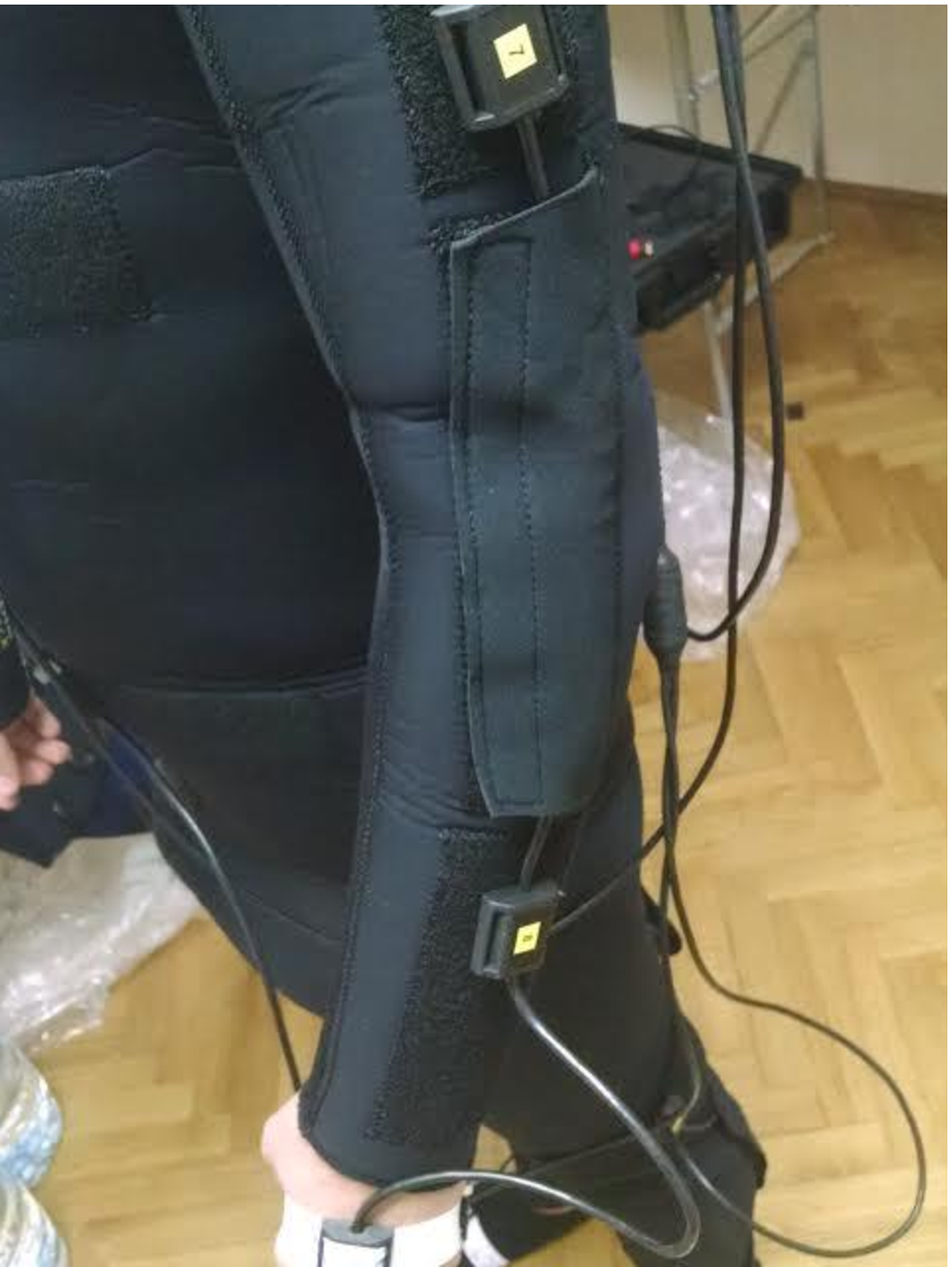}
		\caption{}\label{fig:divernet_suit}
	\end{subfigure}
	\begin{subfigure}[b]{0.98\linewidth}
		\centering
		\includegraphics[height=4.2cm]{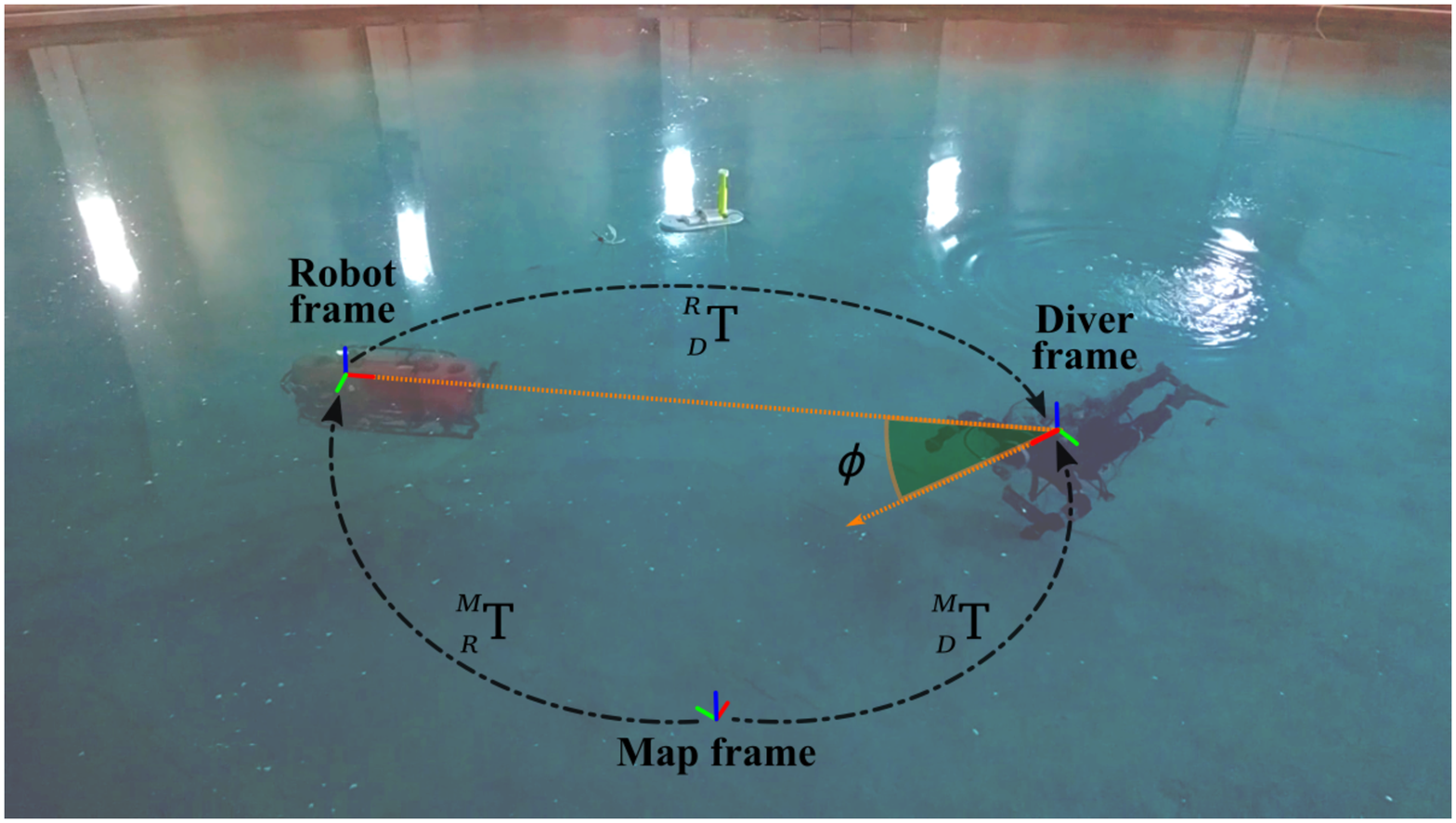}
		\caption{}\label{fig:divernet_heading}
	\end{subfigure}
	\captionsetup{justification=justified}
	\caption{(a) Position of the \emph{DiverNet} 17 IMUs - red dots. (b)\emph{Divernet} suit with sewed in sensors. (c) Computed heading $\phi$, \textit{filtered orientation}, of the diver relative to the AUV.}
	\label{fig:divernet setup}
\end{figure}

During body pose recordings, divers used the \emph{DiverNet}; its hardware, software and data acquisition modules are described in detail in \cite{Goodfellow2015_divernet}. 
In summary, \emph{DiverNet} is a network of 17 Pololu MinIMU-9 Inertial Measurement Units (IMUs) with 9 degrees of freedom (DoFs). 
They are distributed and mounted as shown in Fig.~\ref{fig:divernet_sensor_position},~\ref{fig:divernet_suit}: 3 on each arm and leg, 1 for each shoulder and 1 for the head, torso, and lower back. 
Since it is practically impossible to have the sensors perfectly aligned and firmly in place during tests, a calibration procedure is performed by asking the diver to hold a T-posture and rotating each sensor's coordinate frame to the expected pose. 
Raw and filtered orientation for each sensor is then computed as follows:

\begin{itemize}
	\item \textit{Raw orientation} is acquired based on the magnetometer data and the gravity distribution along each of the accelerometer axes.
	\item \textit{Filtered orientation} is computed by fusing the \textit{raw orientation} with the gyroscope data through a Madgwick-Mahony filter \cite{Madgwick11_IMUGradientDescent}. 
\end{itemize} 

\begin{figure*}[!tb]
	\centering
	\captionsetup{justification=centering}
	
	\includegraphics[width=\linewidth]{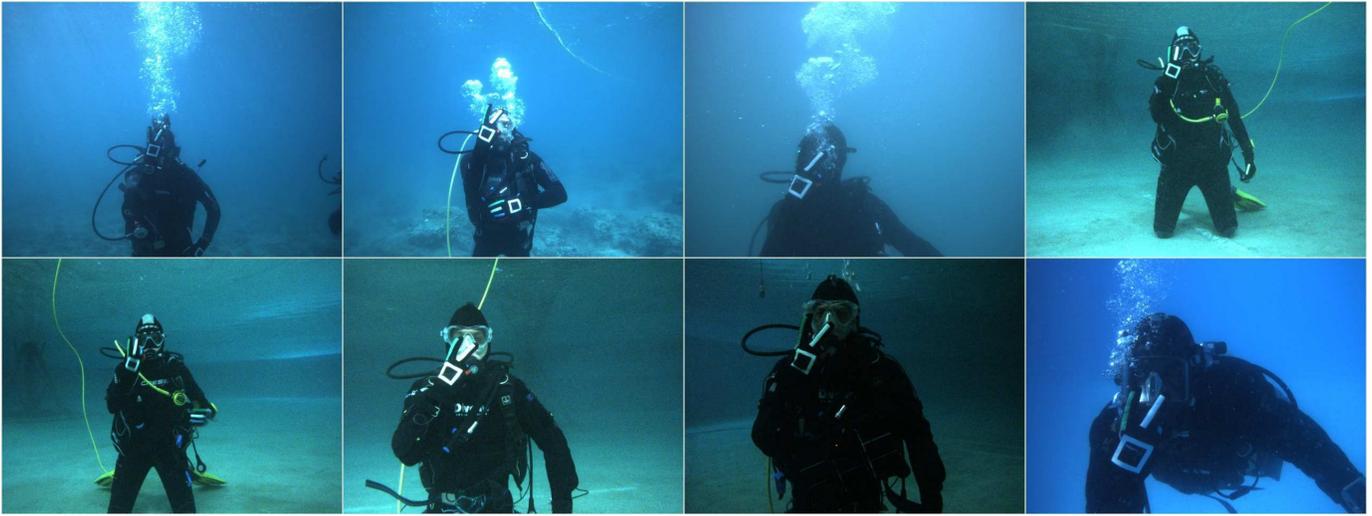}
	\caption{Sample rectified images from each scenario. From left to right and top to bottom: Biograd-A, Biograd-B, Biograd-C, Brodarski-A, Brodarski-B, Brodarski-C, Brodarski-D, Genova-A.}
	\label{fig: scenarios sample images}
\end{figure*}

\begin{table*}[!tb]
	\ra{1.2}
	\centering
	\captionsetup{justification=centering}
	\caption{Description of the different dataset recording scenarios}
	\label{table: gestures_scenarios_description}
	\begin{adjustbox}{max width=.90\textwidth}
		\begin{tabularx}{\textwidth}{@{}XXXXXX@{}}
			\toprule
			\textbf{Location} & \textbf{Scenario} & \textbf{Type} & \textbf{Dynamics} 
			& \textbf{Light \& \newline Distortions} & \textbf{Recording action} \\
			\midrule
			\multirow{3}{*}{\minitab[@{}l]{Biograd na Moru,\\ Croatia}} 
			& Biograd-A & Open sea & No current & Sunlight from \newline the diver's back 
			& Gestures \\ 
			& Biograd-B & Open sea &  No current & Sunlight directly \newline to the diver 
			& Gestures \newline Diver pose \\ 
			& Biograd-C & Open sea &  Strong currents, \newline diver non-static & Fairly dim \newline Haze & Gestures\\
			\midrule 
			\multirow{4}{*}{\minitab[@{}l]{Brodarski Institute,\\ Zagreb, Croatia}}
			& Brodarski-A & Indoor pool &  No current & Bright 
			& Gestures \newline Diver pose \\
			& Brodarski-B & Indoor pool &  No current & Bright \newline Blur
			& Gestures \newline Diver pose \\ 
			& Brodarski-C & Indoor pool &  No current & Fairly dim
			& Gestures \\
			& Brodarski-D & Indoor pool &  No current & Strongly dim
			& Gestures \\
			\midrule
			\multirow{1}{*}{Genova, Italy} 
			& Genova-A & Outdoor pool & Diver non-static & Fairly dim \newline Color absorption
			& Gestures \\ 
			\bottomrule
			
		\end{tabularx}
	\end{adjustbox}
\end{table*}

For data collection, all IMU sensors operate with maximum sensitivity i.e., accelerometer $\pm \SI{16}{g}$,
magnetometer $\pm \SI{1.2}{mT}$, and gyroscope $\pm \SI{2000}{deg/s}$. 
Values are recorded at \SI{50}{Hz} through an optic fiber connecting the \emph{DiverNet} data acquisition unit to an on-land workstation. 
 
The \emph{filtered orientation} consists of absolute values, globally referenced to the map frame \textbf{M}; thus, it is computed in the BUDDY-AUV frame \textbf{R} through the transformation ${}^R_M$\textbf{T}. 
The orientation comes from the average input of the torso and lower back IMU, and its angle in the XY plane, denoted as the \textit{heading} $\phi$, it's the one reported in this dataset (see Fig.~\ref{fig:divernet_heading}). 
As in the EU-FP7 CADDY trials, the data can be exploited to obtain the diver's swimming direction (\emph{heading}) and test tracking algorithms with the AUV based on stereo imagery.

\section{Dataset}
\label{sec:dataset}

\subsection{Data collection}
\label{sec:data_collection}

The recordings for both the underwater gesture and diver pose database took place in 3 different locations in open sea, indoor and outdoor pools. 
Respectively in Biograd na Moru and Brodarski Institute, Croatia, and Genova, Italy. 
Then, the collected data was further divided into 8 different \emph{scenarios} according to the type of conditions that have an impact on the image quality. 
Underwater gestures were recorded in all of them, whereas diver pose/heading only on 3 of them.

Table~\ref{table: gestures_scenarios_description} presents a description of these scenarios, including their dynamic and environmental properties. 
Dynamics refer to the relative motion between AUV and the diver caused by currents or the diver swimming; environmental characteristics mainly cover type of illumination (bright, fairly or strongly dim) and other distortions i.e., blur, haze and color absorption (high blue component). 
Figure~\ref{fig: scenarios sample images} shows a sample image from each setting. All of the described data and software tools for its analysis and visualization are hosted at \url{http://caddy-underwater-datasets.ge.issia.cnr.it/}

\subsection{Underwater gestures}
\label{sec:uw_gestures}

\begin{figure}[!b]
	\centering
	\captionsetup{justification=centering}
	\includegraphics[width=\linewidth]{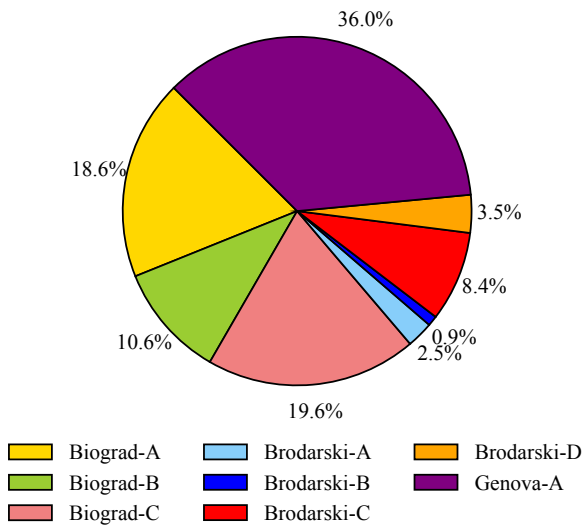}
	\caption{Samples distribution per scenario (True Positives)}
	\label{fig:scenario_distribution}
\end{figure}

The underwater gesture database is a collection of annotated rectified stereo images from divers using the CADDIAN gesture based language~\cite{Chiarella15_CADDIAN}. 
It is important to mention that the number of samples and the class distribution in each scenario varies significantly (see Fig.~\ref{fig:scenario_distribution},\ref{fig:class_distribution},\ref{fig:classes_per_scenario}). 
This is due to the fact that recordings were done at different development stages of the EU FP7 CADDY project.

Biograd-A and B, and Genova-A trials were done mainly for data collection, hence their high number of samples; the rest of the data was collected during test experiments and real diver missions. 
However, since all of these scenarios have different environmental conditions and image quality levels, they are useful to:
\begin{itemize}
	\item Test algorithms and/or image features robustness across different unseen settings i.e., image distortions.
	\item Investigate which underwater image distortions have greater impact on classification methods.
	\item Balance the number of training samples used per scenario to achieve better performance.
	\item Find approaches that fuse 2D and 3D information from the stereo pairs to boost performance.
	\item Test not only object classification methods but also object detectors i.e., locate the diver's hand.   
\end{itemize}

For reference, we also mention that the diver's gloves have a \SI{2.5}{cm} radius circle and a \SI{5}{cm} square in the forehand and backhand respectively; both with a \SI{1}{cm} white border. 
Likewise, each finger has a color stripe, HSV colors and fingers are associated as follows: $\{index:(155,90,50),middle:(0,0,100),ring:(200,90,90),little:(0,100,100),thumb:(70,60,100)\}$. 
The main goal was to provide a defined texture for the diver's hands (the target object) to help classification and disparity calculation and, at the same time, different colors in the fingers help studying underwater color absorption. 

\subsubsection{Data description}
\label{sec:uw_data_description}
\phantom{This text will be invisible}\hfill

\begin{figure}[!b]
	\centering
	\captionsetup{justification=centering}
	\includegraphics[width=\linewidth]{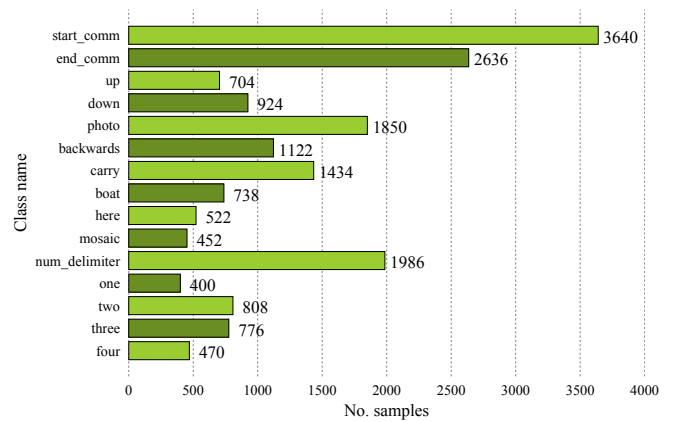}
	\caption{Distribution of gesture classes}
	\label{fig:class_distribution}
\end{figure}

From all the mentioned scenarios in the previous section, \textbf{9231 annotated stereo pairs were gathered for 15 classes (gesture types) i.e., 18462 total samples}. 
Fig.~\ref{fig:class_distribution} shows the distribution of samples per class. 
It is evident that the number of samples is considerably higher for some classes, but since the data was acquired from real diver missions, it is representative of the CADDIAN language distribution in the same way we use some words more frequently than others in our daily speech.
 
\begin{figure}[!b]
	\centering
	\captionsetup{justification=centering}
	\includegraphics[width=\linewidth]{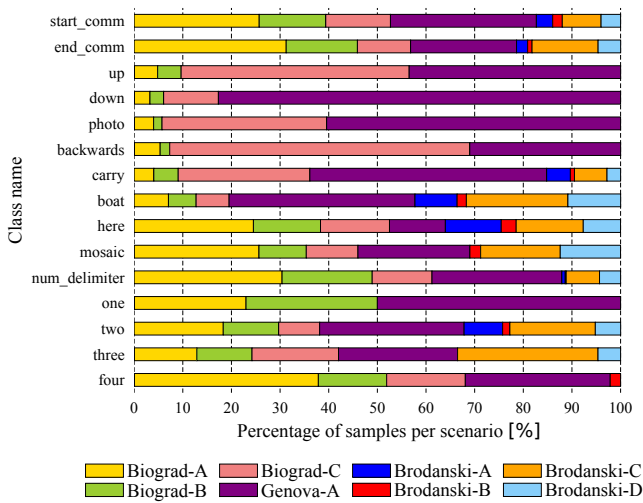}
	\caption{Distribution of classes per scenario}
	\label{fig:classes_per_scenario}
\end{figure}

Due to this and the different class distribution each recording scenario has, we provide Fig.~\ref{fig:scenario_distribution},~\ref{fig:class_distribution} and~\ref{fig:classes_per_scenario} for the user to decide how to build their training/test sets to suit their application; for example, follow the original distribution or build a balanced data set.   
Likewise, we include \textbf{7190 true negative stereo pairs (14380 samples)} that contain background scenery and diver without gesturing; these follow the same distribution per scenario as the true positives in Fig.~\ref{fig:scenario_distribution}.

\subsubsection{Data parsing}
\label{sec:uw_data_parsing}
\phantom{This text will be invisible}\hfill

\begin{table*}[!tb]
	\ra{1.2}
	\centering
	\captionsetup{justification=centering}
	\caption{Header from underwater gestures database \emph{*.csv} file. Row 0 describes the data fields, Row 1 and Row 2 have an example of a true positive and a true negative sample respectively.}
	\label{table:uw_gestures_file_header}
	\begin{adjustbox}{max width=.90\textwidth}
	\begin{tabularx}{\textwidth}{@{}smbbmmbb@{}}
		\toprule
		&\textbf{scenario} & \textbf{stereo left} & \textbf{stereo right} & 
		\textbf{label name} & \textbf{label id} & 
		\textbf{roi left} & \textbf{roi right} \\
		\midrule
		
		0 &
		Recording scenario name &  Filepath to stereo left image  & 
		Filepath to stereo right image & 
		String class name & Integer class ID & 
		Array: \lbrack top corner X, top corner Y, width, height\rbrack & 
		Array: \lbrack top corner X, top corner Y, width, height\rbrack \\ 
		
		1 &
		biograd-A &  /biograd-A \textunderscore00003\textunderscore left.jpg & 
		/biograd-A \textunderscore00003\textunderscore right.jpg & 
		boat & 7  & 
		\lbrack 231,231,62,83\rbrack & \lbrack 152,231,62,83\rbrack \\
		
		2 &
		genova-A &  /genova-A \textunderscore00012\textunderscore left.jpg & 
		/genova-A \textunderscore00003\textunderscore right.jpg & 
		true\textunderscore neg & -1  & 
		\emph{NaN} & \emph{NaN} \\ 
		\bottomrule
		
	\end{tabularx}
	\end{adjustbox}
	
\end{table*}

All of these data is compiled in tabular form in \textit{*.csv} files as they do not require any extra software to be handled and most data analysis packages have built-in methods to process them. 
One file contains the true positives data and other the true negatives. 
Table~\ref{table:uw_gestures_file_header} shows the header/column fields in these files; row with index 0 contains a brief description of the field data, row index 1 shows an example of how a true positive image is referenced and row index 2 an example from a true negative. 
An explanation of these fields is also given below:

\begin{itemize}
	\item \textit{Scenario:} Name corresponding to a location that encompasses particular settings affecting the quality of the image according to Table~\ref{table: gestures_scenarios_description}.
	
	\item \textit{Stereo left/right:} String to the path where the image is saved. The file basename also indicates the scenario from which it is taken, a sequence number and the stereo image it corresponds; in this order e.g., \textit{biograd-A\textunderscore00123\textunderscore left.jpg}.
	
	\item \textit{Label name:} String that identifies the gesture class.
	
	\item \textit{Label id:} Integer that identifies the gesture class.
	
	\item \textit{Roi left/right:} Arrays that describe the regions of interest in the left/right image i.e., where the hand-gesture is located. Each array element is separated by a comma. When 2 instances of the target object are present in the image, each array is separated by a semicolon (this is only true for the \emph{mosaic} gesture). 
	
\end{itemize}

To augment and complement this database, we added 4 different types of distortions to the original stereo-pairs: \textit{blur, contrast reduction, channel noise} and \textit{compression}. These standard distortions are the most commonly present while collecting and transmitting underwater data.  
The user can utilize these \emph{synthetic} images or choose to apply these or other distortions themselves; nonetheless, they are briefly described here as some extra header/column files are reserved to describe them. 
Then, the user can take advantage of this format, the directory structure presented in Fig.~\ref{fig:uw_gestures_dirtree} and the provided software scripts to log their own synthetic images. 

Table~\ref{table:uw_gestures_file_header_augementation} shows the additional columns used to reference synthetic images; the \emph{param} columns store key values of the applied distortions e.g., for blur, \emph{param 1} represents kernel size and for compression, \emph{param 1} and \emph{param 2} refer to the compression scheme and quality level.

\begin{table}[tb]
	\small
	\ra{1.2}
	\centering
	\captionsetup{justification=centering}
	\caption{Additional header/column fields for database augmentation. Row-0 shows the column values for \emph{raw/non-synthetic} image, and Row-1 to 3 for distorted (synthetic) images}
	\label{table:uw_gestures_file_header_augementation}
	\begin{tabularx}{\linewidth}{@{}scXXX@{}}
		\toprule
		&\textbf{synthetic} & \textbf{distortion} &
		\textbf{param 1} & \textbf{param 2} \\
		\midrule	
		0 & 0 &  \emph{NaN} &  \emph{NaN} & \emph{NaN} \\
		1 & 1 &  blur &  13 & \emph{NaN} \\
		2 & 1 &  low contrast  & 0.6  & \emph{NaN} \\
		3 & 1 &  compression  & jpeg  &  65 \\
		\bottomrule

	\end{tabularx}	
\end{table}

\subsubsection{Database directory and provided software}
\label{sec:uw_database directory}
\phantom{This text will be invisible}\hfill

\begin{figure}[!tb]
	\centering
	\captionsetup{justification=centering}
	\includegraphics[width=\linewidth]{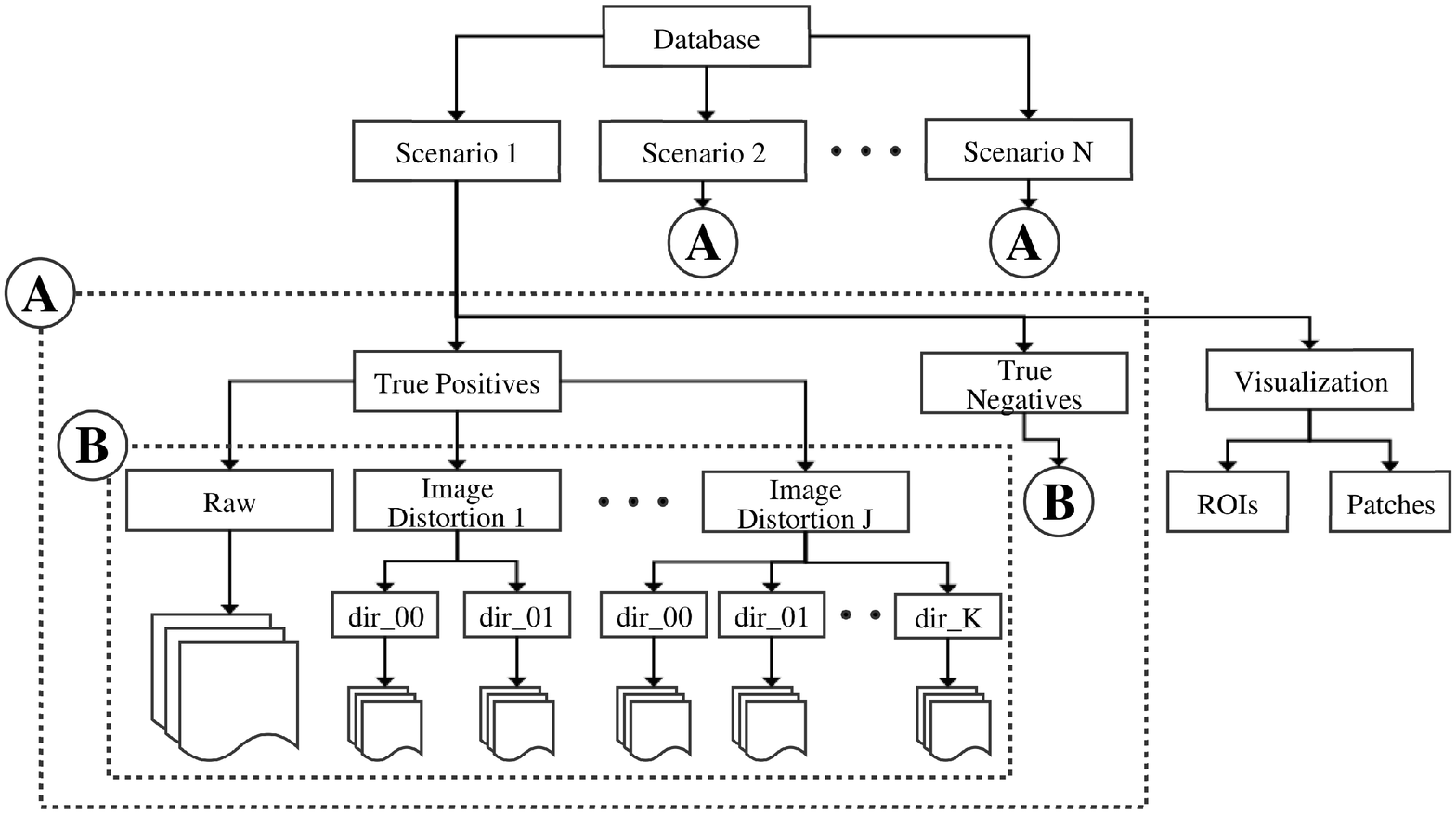}
	\caption{Underwater gestures database directory tree}
	\label{fig:uw_gestures_dirtree}
	
\end{figure}

The provided image files follow the directory structure shown in Fig~\ref{fig:uw_gestures_dirtree}.
As stated in section~\ref{sec:data_collection}, the dataset is first divided by scenarios; which contain a \emph{true positives} and a \emph{true negatives} folder. 
Then, each of these folders contain a \emph{raw} directory with all the original rectified stereo pairs, plus a directory for each image distortion applied to them. 
Since we can apply a particular image distortion with different parameters, a subdirectory named \emph{dir\textunderscore$\#\#$} is created for each different set of parameters used. 
The correspondence between these subdirectories and the distortion parameters can be checked in the database tabular file (see Tables~\ref{table:uw_gestures_file_header},\ref{table:uw_gestures_file_header_augementation}). 
Finally, an extra folder named \textit{visualization} is available for each scenario, where images with highlighted region of interest (ROIs) or hand gesture patches are saved.

In summary, we provide the following tools/scripts for parsing and visualizing the described data, and files the user can utilize as sensor reference. Their usage is explained in \url{http://caddy-underwater-datasets.ge.issia.cnr.it/}.
\begin{itemize}
	\item Parse and visualization scripts to:
	\begin{itemize}
		\item parse by label ID, label name and/or recording scenario.
		\item visualize and/or extract region of interests (ROIs).
		\item apply the mentioned image distortions with user defined parameters.
	\end{itemize}
	\item Camera intrinsic calibration files
	\item Software tools to calibrate underwater cameras with flat-glass panels (\emph{CamOdCal}$+$\emph{Pinax}).
\end{itemize}

\subsection{Diver pose estimation}
\label{sec:diver_pose_estimation}

The diver pose/heading database is as well a collection of annotated rectified stereo images extracted from video sequences showing the diver free-swimming; each stereo pair is associated with a \emph{diver heading} as explained in Section~\ref{sec:DiverNet}. 
In the CADDY project, the primary objective was to track the diver and position the AUV in front as for the diver to always face the camera. 
In this way, the AUV can monitor the diver's activities and communicate through gestures or the underwater tablet shown in Fig.~\ref{fig:buddy-auv}.  
 
Thus, the dataset can be used to test human pose estimation, segmentation or scene geometry understanding methods in this particular context e.g., our work in~\cite{GChavez2017_LSTMDiverPose}.
For \emph{one-shot} or \emph{frame-by-frame} algorithms we offer the rectified stereo pairs while for methods that consider the input history (i.e., diver's previous movements) we provide a sequence number explained in the next section (see Table~\ref{table:uw_dp_file_header}). Data was collected from scenarios Biograd-B, Brodarski-A and Brodarski-C.

\subsubsection{Data description}
\label{sec:dp_data_description}
\phantom{This text will be invisible}\hfill

To collect the data, divers were asked to perform three tasks in front of the AUV: (1) turn \SI{360}{deg} horizontally (chest pointing downwards, to the floor) and (2) vertically, clockwise and anticlockwise, and (3) swim freely. 
For the latter, the AUV was operated manually to follow the diver. 
In total, \textbf{12708 rectified stereo pair images} are provided from which 3D representations can be computed as well.

The collected measurements have passed through a noise (median) filter with a buffer size 5, and an offset correction step (sensor bias) done manually before each test.
As mentioned, $\phi$ is the diver's angle in the XY plane relative to the AUV (frame \textbf{R}) and \SI{0}{deg} is defined when the diver is facing the camera (see Fig~\ref{fig:divernet_heading}). Hence, the range of values go from \SI{-180}{deg} to \SI{180}{deg}. 

\subsubsection{Data parsing}
\label{sec:dp_data_parsing}
\phantom{This text will be invisible}\hfill

This dataset is also presented in tabular \emph{*.csv} form as in Table~\ref{table:uw_dp_file_header}. The explanation of its headers is as follows:
\begin{itemize}
	\item \emph{Scenario:} Name corresponding to recording location and specific settings as in Table~\ref{table: gestures_scenarios_description}.
	\item \emph{Sequence:} Integer that identifies the sequence to which the stereo pair belongs. An image only belongs to a sequence if its from the same scenario and forms part of set continuous in time.
	\item \textit{Stereo left/right:} c.f. Table~\ref{table:uw_gestures_file_header}.
	\item \emph{Heading:} Float number in degrees that indicates the diver heading.
\end{itemize}

\begin{table}[!tb]
	\small
	\ra{1.2}
	\centering
	\captionsetup{justification=centering}
	\caption{Header/column fields for diver heading database. Row 0 describes the data fields, Row 1 and Row 2 show examples}
	\label{table:uw_dp_file_header}
	\begin{tabularx}{\linewidth}{@{}slmmmm@{}}
		\toprule
		&\textbf{scenario} & \textbf{sequence} &
		\textbf{stereo left} & \textbf{stereo right}  & \textbf{heading}\\
		\midrule	
		0 & Scenario Name &  Integer ID &  c.f. Table~\ref{table:uw_gestures_file_header} 
		& c.f. Table~\ref{table:uw_gestures_file_header} & Float (\SI{}{\deg}) \\
		1 & Brodarski-B &  0 & Table~\ref{table:uw_gestures_file_header} & 
		Table~\ref{table:uw_gestures_file_header} & 31.42 \\
		2 & Biograd-A &  3 & Table~\ref{table:uw_gestures_file_header}& 
		Table~\ref{table:uw_gestures_file_header} & -74.51 \\
		\bottomrule
		
	\end{tabularx}	
\end{table}

\subsubsection{Database directory and provided software}
\label{sec:dp_database directory}
\phantom{This text will be invisible}\hfill

The provided video sequences are just split into directories for each scenario, as the first level of the directory structure in Fig.~\ref{fig:uw_gestures_dirtree}. We also offer software tools to: 
\begin{itemize}
	\item Extract a stereo pair given a scenario name, a sequence or a combination of both.
	\item Extract all stereo pairs associated with a range of \emph{heading} values.
	\item Output a sequence as video file for visualization purposes.
\end{itemize}

\section*{Acknowledgements}
	The research leading to the presented results was supported in part by the European Community's Seventh Framework Programme under grant agreement n.~611373 ``Cognitive Autonomous Diving Buddy (CADDY)''.
	This work was also possible thanks to the enthusiasm, collaboration and patience of all the divers that participated in the CADDY project, specially Pavel Ankon and Maša Frleta Valić. Likewise, thanks to Maria Elena Chavez Cereceda for her assistance and time in verifying the integrity of the data here presented.

\bibliographystyle{IEEEtran}
\bibliography{./bibliography}

\end{document}